\documentclass[runningheads]{llncs}
\usepackage[T1]{fontenc}
\usepackage{graphicx}
\usepackage{amsmath}
\usepackage{tabularx, makecell, booktabs} 
\usepackage{multirow}
\usepackage{amsmath}
\usepackage{amssymb}
\begin{document}
\title{\textit{SEBVS}: Synthetic Event-based Visual Servoing for Robot Navigation and Manipulation }
\titlerunning{Synthetic Event-based Visual Servoing}
%
\author{Krishna Vinod\thanks{These authors contributed equally to this work.}\inst{1}\and
Prithvi Jai Ramesh\protect\footnotemark[1] \inst{1}\and Pavan Kumar B N\inst{2} \and
Bharatesh Chakravarthi\inst{1}}
\authorrunning{K. Vinod et al.}
%
\institute{Arizona State University, Tempe, AZ 85281, USA \\ \and
Indian Institute of Information Technology, Sri City, Chittoor, AP, India \\
}
\maketitle              
\begin{abstract}

Event cameras offer microsecond latency, high dynamic range, and low power consumption, making them ideal for real-time robotic perception under challenging conditions such as motion blur, occlusion, and illumination changes. However, despite their advantages, synthetic event-based vision remains largely unexplored in mainstream robotics simulators. This lack of simulation setup hinders the evaluation of event-driven approaches for robotic manipulation and navigation tasks.
This work presents an open-source, user-friendly v2e robotics operating system (ROS) package for Gazebo simulation that enables seamless event stream generation from RGB camera feeds. The package is used to investigate \textit{event-based robotic policies} (\textit{ERP}) for real-time navigation and manipulation. Two representative scenarios are evaluated: ($1$) object following with a mobile robot and ($2$) object detection and grasping with a robotic manipulator. Transformer-based \textit{ERP}s are trained by behavior cloning and compared to RGB-based counterparts under various operating conditions.
Experimental results show that event-guided policies consistently deliver competitive advantages. The results highlight the potential of event-driven perception to improve real-time robotic navigation and manipulation, providing a foundation for broader integration of event cameras into robotic policy learning.
The GitHub repo for dataset and code: \url{https://eventbasedvision.github.io/SEBVS/}

\end{abstract}

\section{Introduction}
\label{sec:intro}
Robots are increasingly deployed in environments that are fast-changing, cluttered, and often extreme, where perception and control must operate with low latency and high robustness. Conventional frame-based vision systems struggle under such conditions. They suffer from motion blur, limited dynamic range, and latency bottlenecks, leading to poor performance in high-speed navigation, precise manipulation, and closed-loop control.

\begin{figure}[h]
\centering
\includegraphics[width=0.95\linewidth]{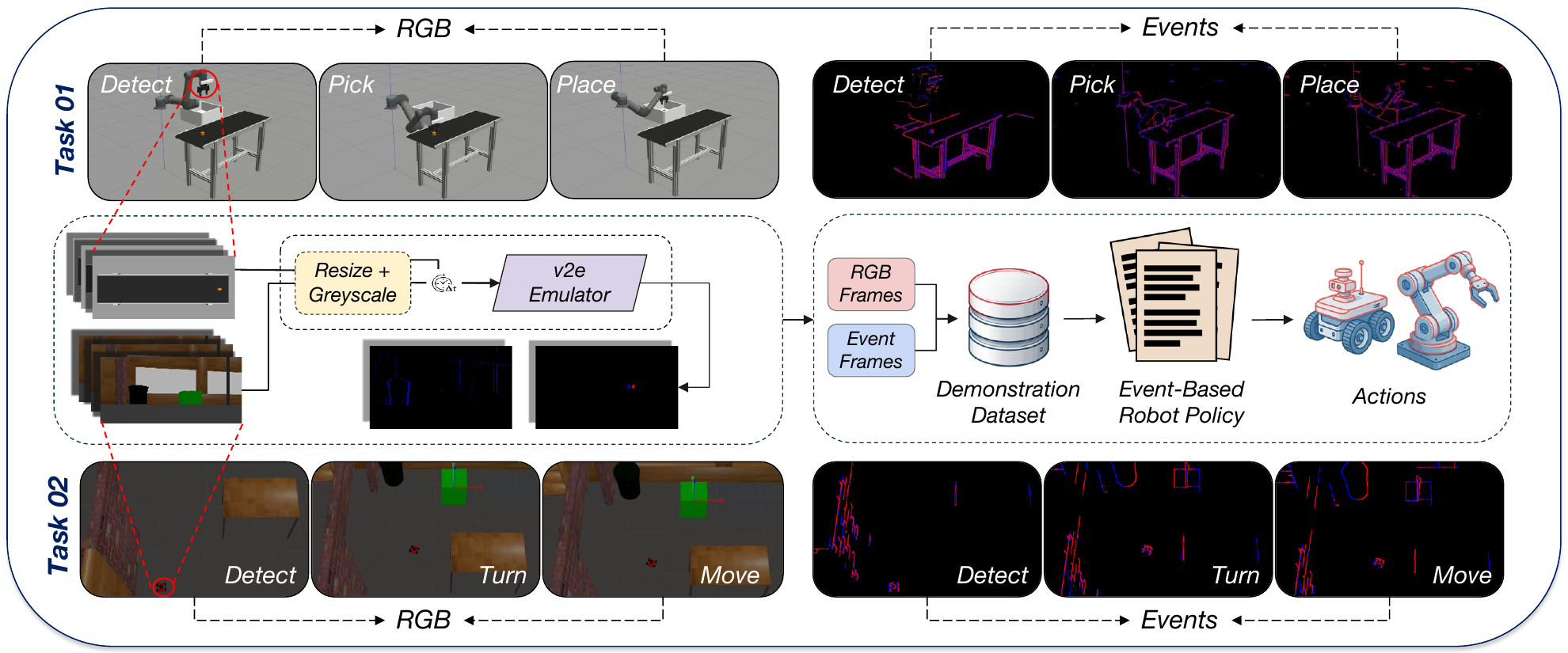}
\caption{Overview of \textbf{s}ynthetic \textbf{e}vent-\textbf{b}ased \textbf{v}isual \textbf{s}ervoing (\textit{SEBVS}) framework for manipulation (\textit{Task01}) and navigation (\textit{Task02}).}
\label{fig01}
\end{figure}

Event cameras \cite{chakravarthi2024recent,Gallego2022Survey}, also known as neuromorphic or dynamic vision sensors (DVS), offer a compelling alternative. Unlike conventional cameras that capture frames at fixed rates, event cameras asynchronously report per-pixel brightness changes with microsecond latency, very high dynamic range, and zero motion blur. These characteristics make them ideally suited for a wide range of applications, ranging from traffic monitoring \cite{verma2024etram,chakravarthi2023event}, autonomous driving \cite{aliminati2024sevd,tan2025real}, pose estimation \cite{chanda2025sepose}, human computer interaction, and surveillance, for time critical robotic tasks, including high speed navigation, agile manipulation, and visual servoing under extreme lighting \cite{shravan2023innovative}.

Despite their promise, the adoption of event-based vision in robotics has been limited, particularly in simulation. There are two primary challenges. First, mainstream robotics simulators, such as Gazebo, lack native event-camera models. Researchers are forced to rely on offline converters such as ESIM  \cite{Rebecq2018ESIM} or v2e \cite{Hu2021v2e} to generate synthetic event streams from rendered RGB video. Although effective for data set creation, these tools are cumbersome, time-consuming, and ill-suited for interactive policy training. Second, the sparse, asynchronous nature of event streams does not align with traditional perception pipelines, which are designed for dense, synchronous RGB frames. Leveraging the rich spatio-temporal structure of event data for robot control requires dedicated learning architectures.

Recent studies have made progress toward closing this gap. \cite{Gehrig2023RVT} introduced recurrent vision transformers (RVT) for event-based object detection, demonstrating that transformer backbones can process event streams efficiently. \cite{Bugueno2025HRL} demonstrated event-camera-driven reinforcement learning controllers for robot navigation, highlighting the feasibility of event-driven policies. However, there remains no simulation-ready, end-to-end framework that allows learning closed-loop event-based robot policies covering both perception and control, directly in a robotics simulator.
To address this gap, we present an integrated framework (see Fig.~\ref{fig01}) \textit{SEBVS}, for event-driven robotic perception and control in simulation. The contributions are summarized below:

\begin{enumerate}
    \item An open-source ROS2/Gazebo package that integrates a real-time event stream via a fine-tuned v2e back-end, enabling low-latency neuromorphic perception without specialized hardware.
    
    \item  A transformer-based architecture that fuses accumulated event frames with RGB images via patch embeddings, positional encoding, and stacked self-attention blocks to produce high-frequency control outputs.
    
    \item Evaluation of two representative tasks: (i) tracking of mobile robot objects (cube) for navigation, and (ii) UR5 with Robotiq $2$F-$85$ gripper for pre-grasp pose prediction for manipulation. Policies are trained via behavior cloning from expert demonstrations, where event data is collected from our open-source package.
    
    \item  Comparison of event-driven policies against RGB-only baselines, measuring tracking accuracy, task success rate, and closed-loop latency.
\end{enumerate}

Experiments demonstrate that event-based robot policies consistently outperform RGB-based baselines, offering faster reaction times and greater robustness to motion blur and lighting variations. These results underscore the practical benefits of neuromorphic sensing for real-time robotic navigation and manipulation and establish a foundation for the broader integration of event cameras into simulation-to-real robot policy learning.

\section{Related Works}
\label{Sec:RelatedWorks}

Event-based vision enables robust perception in high-speed, dynamic environments commonly encountered in robotics. This section reviews the foundations of event-based sensing, simulation frameworks using DVS data, and learning-based methods for event-driven control in robotic tasks.

\subsection{Event-based Vision for  Robotics} 
\label{Sec:2.1}
Event-based vision has progressed from niche demonstrations to a key enabler of high-speed robotic perception. Surveys such as Zheng et al. \cite{zheng2023deep} consolidate deep learning methods for spike-based data and highlight open challenges in robotic integration.
In robotic navigation, event cameras have been deployed in complex and cluttered environments. Zujevs et al. \cite{zujevs2021event} introduced the first DVS navigation dataset for agricultural fields, enabling robust evaluation under severe occlusions. Bugueno-Córdova et al. \cite{Bugueno2025HRL} integrated event cameras with reinforcement learning to develop human-aware navigation policies, demonstrating improved performance in crowded environments. Palinauskas et al. \cite{palinauskas2023generating} coupled ESIM with MuJoCo to generate fully labeled synthetic streams, accelerating policy prototyping.

In robotic manipulation, event-based sensing has enabled rapid grasping and dynamic interaction. Li et al. \cite{li2020event} proposed an event-grasping dataset with a CNN pipeline to localize stable grasps directly in the spike domain, while Huang et al. \cite{huang2022real} demonstrated microsecond-latency grasping of objects moving at speeds exceeding $2$ m/s. Beyond grasping, event cameras have also facilitated high-speed interaction tasks: Ziegler et al. \cite{ziegler2025event} achieved $2$ ms visual feedback for a table-tennis robot, and Vitale et al. \cite{vitale2021event} implemented fully event-driven vision and control on an embedded UAV platform.
Event cameras have additionally supported multi-robot tracking at kilohertz rates \cite{iaboni2021event} and enhanced pedestrian-aware navigation policies \cite{Bugueno2025HRL}. Collectively, these works demonstrate the versatility of asynchronous sensing across ground, aerial, and manipulation tasks. However, most solutions rely on custom pipelines tailored to specific applications. A unified simulation-ready framework for event-driven policy learning, training both navigation and manipulation policies, remains unexplored. By embedding a high-fidelity v2e converter inside Gazebo and benchmarking transformer-based policies across domains, this work aims to address this gap and foster reproducible, large-scale research on event-driven robotic tasks.

\begin{table}[t]
\caption{Event-camera support widely used in robotic simulators.}
\label{table01}
\resizebox{\columnwidth}{!}{%
\begin{tabular}{c|c|c|c}
\toprule 
\textbf{Simulator} & \textbf{Rendering Engine} & \textbf{Native Event Support} & \textbf{Third-party/Plugin Support} \\ \midrule 
Gazebo (Classic/Garden) & OGRE/USD         & No  & \texttt{gazebo\_dvs\_plugin} (Community, outdated) \\ \hline
Isaac Sim               & Omniverse RTX/US & No  & Custom \texttt{RTX} extensions (Closed-source)     \\ \hline
CARLA Sim               & Unreal Engine    & Yes & --                                         \\ \hline
MuJoCo                  & OpenGL           & No  & \texttt{MuJoCo-ESIM}                               \\ \bottomrule 
\end{tabular}%
}
\end{table}

\subsection{Event Camera Simulators for Robotics}
\label{sec:2.2}
Despite the growing adoption of event cameras in robotics, their integration into mainstream simulators remains limited. Accurate simulation is essential for training, benchmarking, and validating event-driven policies at scale, yet most platforms provide no native DVS support and rely on external tools.
\textbf{\textit{Vision-only solutions}}: ESIM \cite{Rebecq2018ESIM} and v2e \cite{Hu2021v2e} remain the most widely used vision-based solutions, converting rendered RGB frames into synthetic events, either on-the-fly (ESIM within Unity) or offline (v2e in any image sequence). More specialized converters, such as DVSVoltmeter \cite{Wong2022Voltmeter} and DVS-Gait \cite{Medina2018DVSGait}, incorporate circuit-level noise or focus on specific dataset generation tasks.
\textbf{\textit{Robotic physics simulators:}} (as summarized in Table~\ref {table01}), none of the major simulation engines provides a maintained, first-class DVS sensor. Gazebo relies on the outdated \texttt{\texttt{gazebo\_dvs\_plugin}}, Isaac Sim depends on closed-source RTX extensions, and MuJoCo integrates event simulation through ESIM. As a result, most practitioners revert to the ESIM/v2e workflow, which is functional but cumbersome for users and computationally expensive for large-scale experiments.

\textbf{\textit{Our Motivation:}} These limitations underscore the need for a user-friendly approach that embeds event generation directly into standard robotic simulators. Such integration would streamline experimentation, eliminate extra conversion steps, and lower the barrier to reproducible event-based robotics research.

\subsection{Event-based Robot Policy Learning}
\label{sec:2.3}
Transformer-based scaling has emerged as a dominant paradigm in modern robot policy learning. RT-1 \cite{brohan2022rt} demonstrated a large multi-task visuomotor model capable of controlling a real robot across hundreds of skills from purely RGB demonstrations. RT-$2$ \cite{zitkovich2023rt} extended this approach with vision language action (VLA) pre-training, transferring web-scale knowledge to physical manipulation. OpenVLA \cite{kim2024openvla} further introduced an open-source foundation for community-driven VLA research. Similar scaling trends are seen in mobile robotics, where vision language navigation (VLN) systems such as LM-Nav \cite{xia2023lmnav} integrate large language models with semantic maps to navigate ground robots via natural-language commands. Despite these advances, most of these architectures remain frame-centric.  

Event-centric architectures aim to close this gap by leveraging the asynchronous nature of event data. Vemprala et al. \cite{vemprala2021evvae} proposed a latent \emph{event} VAE (eVAE) and trained a PPO policy for UAV obstacle avoidance, achieving faster convergence and greater robustness than frame-based baselines. Gehrig et al. \cite{Gehrig2023RVT} introduced RVT for event-based object detection, combining spatial self-attention with temporal recurrence. Li et al. \cite{li2020event} developed spiking-aware CNNs for real-time grasp detection directly from raw events. Reinforcement learning studies have further demonstrated the control potential of event data, from neuromorphic quadrotor maneuvers on embedded SoCs \cite{vitale2021event} to pedestrian-aware navigation policies with dynamic vision sensors \cite{Bugueno2025HRL}.  
While these works highlight the promise of event-driven perception, most remain confined to perception pipelines and lack direct comparisons with RGB-trained policies in closed-loop control. This work investigates the potential of transformer-based policies that operate directly on event frames, enabling end-to-end control for both navigation and manipulation tasks.

\section{The \textit{SEBVS} Framework}
\label{sec:3framework}
This section presents the proposed \textit{SEBVS} framework and its application to learning \textit{event-based robotic policies} (\textit{ERP}) for navigation and manipulation. The framework integrates a ROS2-based v2e emulator with Gazebo to generate event streams in simulation, which are then used to train transformer-based control policies via behavior cloning from expert demonstrations.

\subsection{v2e ROS2 Emulator for Gazebo}
\label{sec:3.1}
To enable event-camera simulation in Gazebo, a lightweight ROS2 package was developed that integrates v2ecore's \texttt{EventEmulator} with standard RGB camera topics. The emulator subscribes to the RGB image stream \texttt{/camera/image\_raw}, performs resizing and grayscale conversion, and forwards the processed images to the \texttt{EventEmulator}. The generated event stream is published on the topic \texttt{/dvs/events}, which can subsequently be accumulated into event frames for downstream processing.

The primary objective of the package is to simplify the collection of event data in robotics simulators, thereby reducing the challenges of simulating event-based robotic tasks. The design emphasizes ease of integration with minimal configuration requirements. It operates efficiently on commodity hardware, such as a laptop with an NVIDIA GPU and CUDA support. Basic ROS knowledge, primarily topic remapping in Gazebo, is sufficient for integration into custom simulation pipelines.

The emulator provides several tunable parameters (see Table~\ref {table02}) that allow fine-grained control over event generation. The contrast thresholds, noise levels, and downsampling factors can be adjusted to match specific simulation requirements. This flexibility makes the package suitable for various event camera research scenarios.
The package has been tested on ROS$2$ Humble with Gazebo Classic. Because the emulator operates by subscribing to camera topics and publishing event topics, it is agnostic to specific ROS$2$ or Gazebo versions. With minor adjustments to topic connections, the same node can be adapted for other ROS-based simulation environments.

\begin{figure}[t]
\centering
\includegraphics[width=0.95\linewidth]{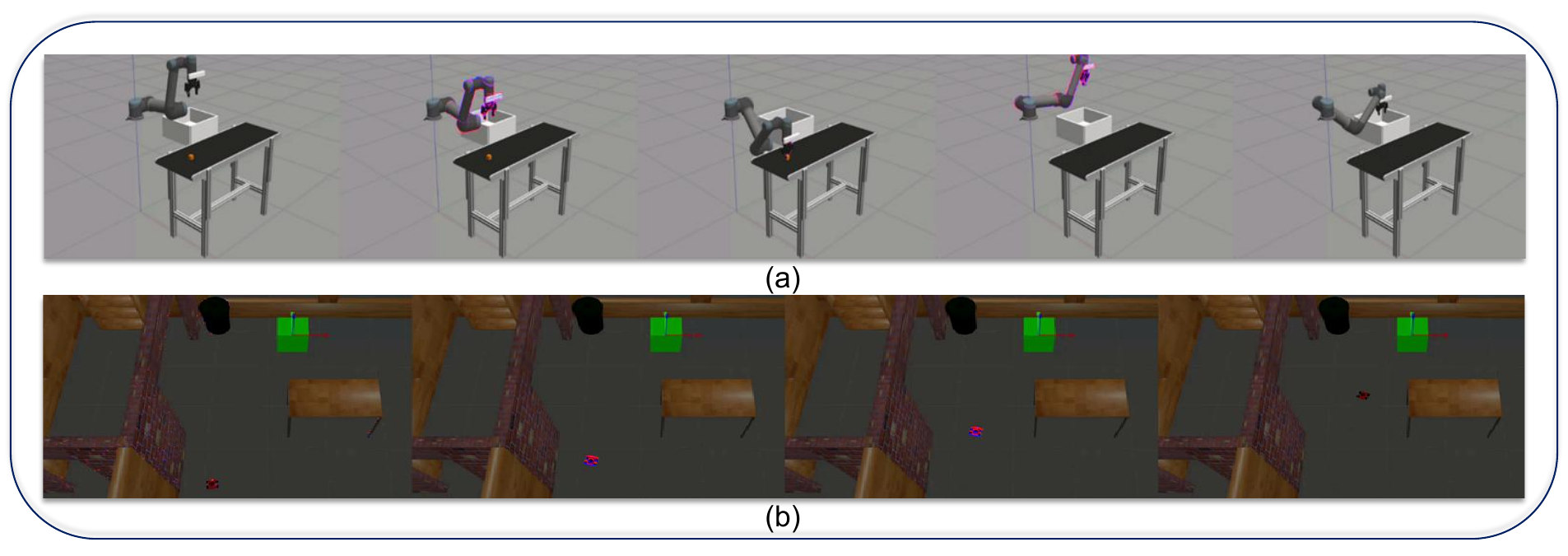}
\caption{Overview of the robotic tasks to experiment with our \textit{ERP}s: a) UR5 pre-grasp pose prediction, and b) Differential bot object tracking. }
\label{fig02}
\end{figure}

\begin{table}[t]
\centering
\caption{Summary of the v2e event-emulator node’s tunable parameters.}
\label{table02}
\resizebox{0.80\columnwidth}{!}{%
\begin{tabular}{c|c|c}
\toprule 
\textbf{Parameter} & \textbf{Default Value} & \textbf{Description}                             \\ \midrule 
\texttt{pos\_thres}         & $0.3$                    & Positive contrast threshold (ON events)          \\ \hline
\texttt{neg\_thres}        & $0.3$                    & Negative contrast threshold (OFF events)         \\ \hline
\texttt{sigma\_thres}       & $0.09$                   & Noise-clamp / sigma threshold                    \\ \hline
\texttt{cutoff\_hz}         & $15.0$                   & High-pass filter cutoff frequency (Hz)           \\ \hline
\texttt{leak\_rate\_hz}     & $0.0$                    & Exponential “leak” reset rate (Hz)               \\ \hline
\texttt{downsample}         & $0.5$                    & Input image downsampling factor (e.g. $0.5$=½ res) \\ \hline
\texttt{blur}               & True                   & Enable $3$×$3$ Gaussian blur before event            \\ \bottomrule
\end{tabular}%
}
\end{table}

\subsection{Event-Driven Control Policies for Navigation and Manipulation}
\label{sec:3.2}
Object detection and tracking are long-standing problems in robotic tasks such as arm manipulation and mobile robot navigation, such as Fig.~\ref{fig02}. While modern vision-based detectors and trackers achieve high accuracy, they remain susceptible to motion blur, latency, and limited dynamic range. The proposed \textit{ERP}s address these limitations by jointly processing synchronized RGB frames and asynchronous event streams to directly output control actions in real time.

At each time step \(t\), the policy receives an observation \(o_t\) consisting of both the RGB image and the accumulated event frame:

\[
\mathbf{o}_t = \bigl(I_t,\,E_t\bigr),
\]
where \(I_t\in\mathbb R^{H\times W\times3}\) is the RGB image, and the event‐frame is
\[
E_t(x,y) \;=\; \sum_{k : t_k \in [t-\Delta t,t]}\! p_k\,\delta(x-x_k,y-y_k),
\]
with \((x_k,y_k,t_k,p_k)\) each event’s pixel location \((x,y)\), timestamp \(t\), and polarity \(p\).  We discretize \(E_t\) into two channels \(E_t^+\) and \(E_t^-\) for \(p_k=+1\) and \(-1\), respectively.

\vspace{1ex}

\textbf{\textit{Task 1: Differential Drive Tracking (\textit{ERPNav)}:}}  
The \textit{ERPNav policy} \(\pi_{\mathrm{nav}}\) maps each observation \(\mathbf{o}_t\) to a \(\mathtt{geometry\_msgs/Twist}\) command, outputting linear \(v_t\) and angular \(\omega_t\) velocities  that reorient the robot towards the target:
\[
\mathbf{u}_t 
= \begin{bmatrix}v_t \\ \omega_t\end{bmatrix}
= \pi_{\mathrm{nav}}(\mathbf{o}_t),
\tag{1}
\]
which controls the robot according to the unicycle kinematics,
\[
\dot{x}_t = v_t \cos\theta_t
\quad
\dot{y}_t = v_t \sin\theta_t
\quad
\dot{\theta}_t = \omega_t
\tag{2}
\]
and the policy is trained via behavior cloning to minimize the mean squared error between predicted and expert commands:

\[
\mathcal{L}_{\rm nav}(\theta)
= \frac{1}{T}\sum_{t=1}^T 
\bigl\|\pi_{\mathrm{nav}}(\mathbf{o}_t)-\mathbf{u}_t^*\bigr\|_2^2.
\tag{3}
\]

\vspace{1ex}

\textbf{\textit{Task 2: UR5 Pre‐Grasp Pose Prediction (ERPArm):}}  
In this task, the \textit{ERPArm policy}\(\pi_{\mathrm{arm}}\) predicts the $6$-DoF pre-grasp end-effector pose required for the UR5 to approach and grasp a target block. From a single initial RGB event observation \(\mathbf{o}_0\), the policy outputs:
\[
\mathbf{p}^*
= \pi_{\mathrm{arm}}(\mathbf{o}_0)
\;\in\;\mathbb{R}^6
\quad
\bigl[
x,\,y,\,z,\,
\phi_{\mathrm{(roll)}},\,
\theta_{\mathrm{(pitch)}},\,
\psi_{\mathrm{(yaw)}}
\bigr]^\top,
\tag{4}
\]
The policy is trained to minimize the mean squared error between the predicted and ground-truth pre-grasp pose:

\[
\mathcal{L}_{\rm arm}(\theta)
= \bigl\|\pi_{\mathrm{arm}}(\mathbf{o}_0)-\mathbf{p}_{\rm true}\bigr\|_2^2.
\tag{5}
\]

By combining the microsecond precision and high dynamic range of event data alongside RGB imagery, the \textit{ERPNav} achieves rapid base reorientation (through eqn.\,(1)–(3)) while \textit{ERPArm} predicts reliable grasp poses for the UR5  (eqn.\,(4)–(5)), mitigating motion blur and latency limitations of frame-only policies.

\begin{figure}[t]
\centering
\includegraphics[width=0.90\linewidth]{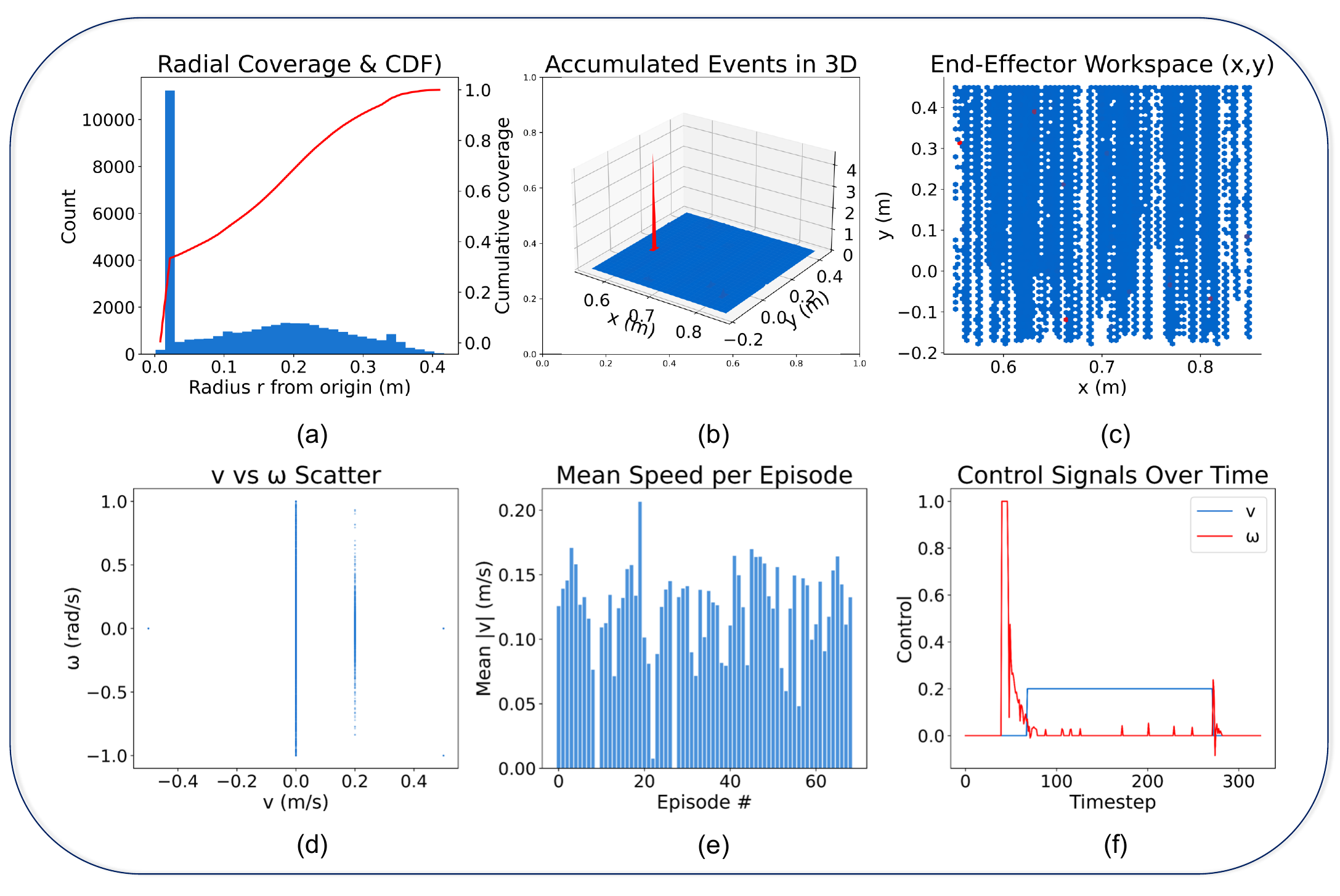}

\caption{Visualization of training datasets for \textit{ERPArm} (top) and \textit{ERPNav} (bottom).  
(a) Radial coverage and cumulative distribution function,   
(b) Accumulated events in 3D (red peak shows the home position),
(c) End-effector workspace distribution, 
(d) Control vector scatter of linear speed vs. yaw rate,  
(e) Mean absolute speed per episode, and  
(f) Example expert command sequence in a mobile robot.  
These plots capture spatial, temporal, and action-space characteristics critical to \textit{ERP} training.}
\label{fig03}
\end{figure}

\subsection {\textit{ERPNav} and \textit{ERPArm} - The Expert Demonstration Dataset}
\label{sec:3.3}
\begin{figure}[t]
\centering
\includegraphics[width=1\linewidth]{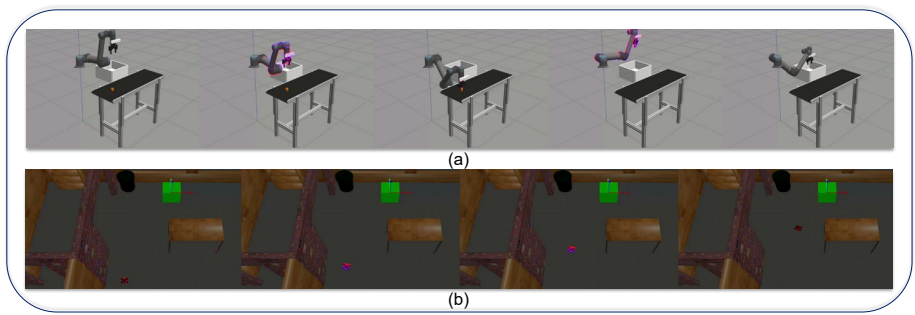}
\caption{Qualitative visuals (camera view) for our \textit{ERPNav} and \textit{ERPArm} datasets}
\label{fig04}
\end{figure}

Two datasets were collected for \textit{ERPNav} and \textit{ERPArm} tasks (see Fig.~\ref{fig04}). In both cases, expert demonstrations were recorded in simulation using YOLO-based perception pipelines, with synchronised RGB and event frames from the v2e emulator, and corresponding expert actions.

\textbf{\textit{ERPNav Dataset:}} The \textit{ERPNav} training set was collected using a four-wheel differential-drive robot equipped with a \(640\times640\) RGB camera and the Gazebo-integrated v2e emulator, which converts each RGB frame into event stacks in real time. The expert policy combined a fine-tuned YOLOv$8$ object detector with a two-axis PID controller. Detected object centroids were fed to the controller, which generated continuous linear-\(x\) and angular-\(z\) velocity commands, forming a ``YOLO+PID'' control loop.  

A custom ROS2 recorder node synchronously logged, at each time step, three data tensors - RGB, event frames (ON/OFF histograms aligned with the RGB image), and velocity actions into an HDF5 file. Additionally, scale-offset parameters mapping raw event coordinates into RGB space were stored.  Seventy episodes (\(\sim\)$65$~min of simulation) were recorded, yielding $32{,}033$ labeled control vectors with balanced distribution (see Fig.~\ref{fig03}). The dataset captures a diverse range of behaviors, from prolonged smooth tracking to sharp corrective maneuvers, providing the transformer policy with precise, time-aligned supervision for low-latency, event-driven navigation. Behavior cloning was applied in a manner consistent with prior large-scale policy training approaches \cite{kim2024openvla,brohan2022rt}, concatenating demonstration episodes for training.

\begin{table}[t]
\caption{\textit{ERPNav} and \textit{ERPArm} - event-based robot-policy datasets for navigation and manipulation tasks}
\label{table03}
\resizebox{\columnwidth}{!}{%
\begin{tabular}{c|c|c|c|c|c}
\toprule
\textbf{Policy} &
  \textbf{Robot} &
  \textbf{Expert Policy} &
  \textbf{Episodes} &
  \textbf{Samples} &
  \textbf{Output Format} \\ \midrule
\textit{ERPNav} &
  \begin{tabular}[c]{@{}c@{}}4-wheel differential-\\ drive mobile robot\end{tabular} &
  \begin{tabular}[c]{@{}c@{}}YOLOv8+\\ PID Controller\end{tabular} &
  \begin{tabular}[c]{@{}c@{}}70 \\ ($\sim$65 minutes)\end{tabular} &
  \begin{tabular}[c]{@{}c@{}}32,033\\ action labels\end{tabular} &
  \begin{tabular}[c]{@{}c@{}}RGB frames, event frames tensors,\\ cmd\_vel (linear x, angular z)\end{tabular} \\ \hline
\textit{ERPArm} &
  \begin{tabular}[c]{@{}c@{}}6-DoF articulated arm \\ with a wrist camera\end{tabular} &
  \begin{tabular}[c]{@{}c@{}}YOLOv8+MoveIt \\ pick-and-place module\end{tabular} &
  \begin{tabular}[c]{@{}c@{}}35 \\ ($\sim$70 minutes)\end{tabular} &
  \begin{tabular}[c]{@{}c@{}}18,417\\ motion way-points\end{tabular} &
  \begin{tabular}[c]{@{}c@{}}RGB frames, event frame tensors,\\ 6-DoF end-effector pose\end{tabular} \\ \bottomrule
\end{tabular}%
}
\end{table}

\textbf{\textit{ERPArm Dataset:}}  
The \textit{ERPArm} dataset was collected using the UR5 equipped with a wrist-mounted RGB camera \texttt{/ee\_camera/image\_raw} and the v2e emulator for event generation. Expert pre-grasp poses were provided by a YOLO-based detector that published object world coordinates \texttt{/object\_world \_pose}. A custom ROS2 node, \texttt{PickPoseRecorder}, synchronously logged the RGB tensor (\(H\times W\times 3\)), a two-channel event tensor (\(H\times W\times 2\)) counting ON/OFF events since the last flush, and a 6-DoF pre-grasp pose vector \((x, \;y, \;z, \;roll, \;pitch,  \;yaw)\). (see Table~\ref{table03}) Thirty-five episodes were recorded, each containing 1{,}000–2{,}000 synchronized frames. To ensure safety, any object with \(y > 0.449\)~m in world coordinates was assigned a predefined \texttt{HOME\_POSE}, preventing the arm from moving to invalid positions when objects left the workspace. The dataset includes demonstrations in both single-object and cluttered multi-object scenarios. These RGB–event–pose triplets enable the \textit{ERPArm} policy to learn a direct, one-shot mapping from perception to pre-grasp configuration, supporting fast and robust grasping in dynamic scenes.

\subsection{Transformer-based Policy Learning for Event-driven Control}
\label{sec:3.4}
\begin{figure}[t]
\centering
\includegraphics[width=1\linewidth]{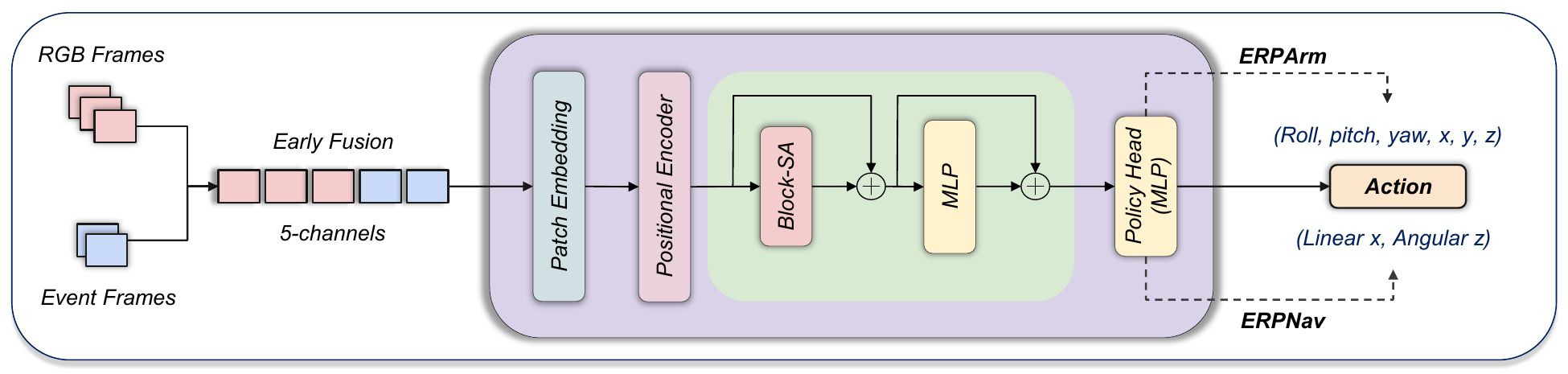}
\caption{\textit{ERP} architecture used for both tasks. RGB frames ($3$ channels) are early-fused with ON/OFF event channel ($2$ channels) to form a $5$-channel input that is tokenised by a patch-embedding layer, enriched with positional encodings, and processed by a lightweight Transformer encoder (self-attention block+MLP, each with residual connections). A task-specific policy head maps the class token to actions: linear-\(x\) and angular-\(z\) velocities for \textit{ERPNav} (mobile robot) or $6$-DoF pre-grasp-pose command for \textit{ERPArm} (manipulator).}
\label{fig05}
\end{figure}

The event-based robot policies (\textit{ERPNav} and \textit{ERPArm}) are end-to-end control networks that jointly process RGB and event frames to generate task-specific control outputs. Both policies share a common backbone architecture (see Fig.~\ref{fig05}), inspired by the vision transformer (ViT)~\cite{dosovitskiy2020image}, while remaining compact enough for real-time inference.
At each time step, the RGB frame (three channels) is concatenated with its two event-count maps (ON and OFF polarities) to form a five-channel input tensor. This early fusion approach combines appearance information with microsecond-scale motion cues at the input stage. The tensor is then divided into non-overlapping \(16\times16\) patches via patch embedding. Each patch is flattened and projected to a $64$-dimensional embedding, yielding a sequence of $1{,}600$ patch tokens. A learnable class token is prepended, and fixed sinusoidal positional encodings are added to preserve spatial information after flattening.
The resulting token sequence is processed by a lightweight Transformer encoder consisting of a four-head multi-head self-attention layer followed by a $256$-unit feed-forward block. Residual connections, layer normalization, and dropout ($0.1$) are applied after each sub-layer, following the ViT and RVT~\cite{Gehrig2023RVT} formulation while keeping the parameter count modest. The updated class token serves as a global spatio-temporal summary of the fused RGB–event observation.
Finally, the class token is passed to a task-specific MLP policy head. For \textit{ERPNav}, the \texttt{PolicyHeadNav} outputs a two-dimensional action \([v_t, \omega_t]\) corresponding to forward linear velocity and yaw rate. For \textit{ERPArm}, the \texttt{PolicyHeadArm} outputs a $6$-DoF end-effector pose \((x,\;y,\;z,\;roll,\;pitch,\;yaw)\). This shared architecture ensures efficient learning across navigation and manipulation tasks while enabling real-time inference.

\section{Experiments}
\label{sec:experiments}
This section presents the \textit{SEBVS} framework across two simulated tasks, outlining the setup, training methods, and evaluation metrics. Policy variants using RGB, Event-only, and fused inputs are compared for training and inference performance.

\subsection{Workspace and Experimental Setup}
\label{sec:4.1}
To evaluate the potential of event-based imitation learning (IL) policies for robotic navigation and manipulation, two simulated tasks were designed. The first task involves a custom four-wheel differential-drive robot operating in a Gazebo environment that simulates both indoor and outdoor settings. In this setup, the \emph{ERPNav} policy is trained to generate control commands for tracking a moving cube. 
The second task focuses on arm manipulation using a Gazebo-based simulation of a UR5 manipulator equipped with a Robotiq 2F-85 gripper. A wrist-mounted RGB camera (640\(\times\)640\,px) captures synchronous visual input, while a co-located v2e emulator generates event streams from the same viewpoint~\cite{Gallego2022Survey}. Target objects are placed on a conveyor belt, which supports both static and moving modes. In both tasks, a cube is used as the object of interest.

The \emph{ERPArm} network ingests the wrist-camera data and regresses a 6-DoF pre-grasp pose (position+orientation) hovering above the target object. A separate module performs downstream grasp execution and placement; here, we focus solely on the accuracy and latency of the pre-grasp pose prediction.
For testing, we train three variants of both \emph{ERPNav} and \emph{ERPArm}. The first \emph{RGB-only} variant uses only RGB frames as input, mirroring the standard behavior-cloning policy in robotics. The second \emph{RGB+Event} variant fuses RGB and event frames using early fusion (as described in Section.~\ref{sec:3.4}) before passing the combined data to the transformer blocks. The third \emph{Event-only} variant relies exclusively on event-frame data to evaluate how the IL policy performs with event inputs alone. Training all three variants for each task enables a comprehensive analysis of the \textit{ERP} policies’ potential.

\begin{figure}[t]
\centering
\includegraphics[width=0.95\linewidth]{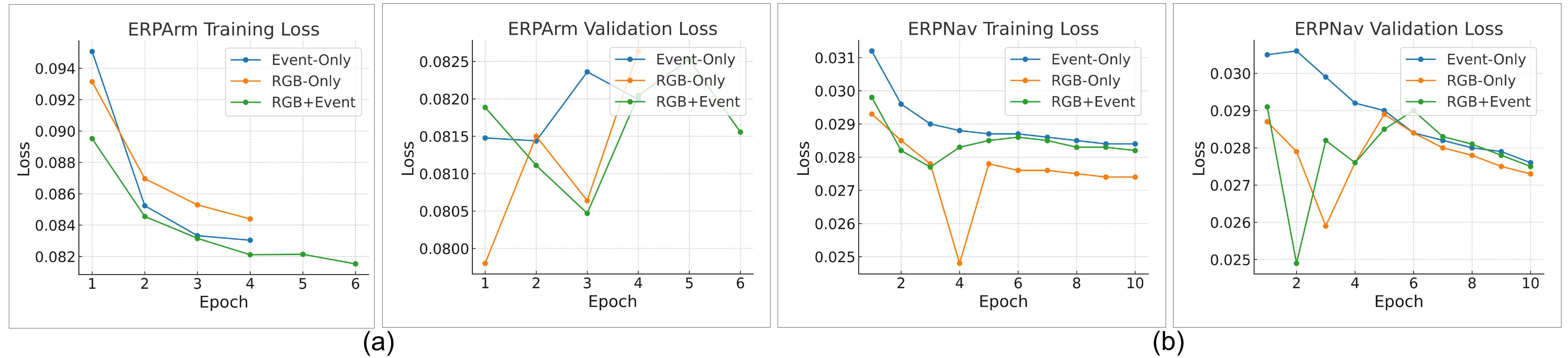}
\caption{Training- and validation-loss curves for the \textit{ERPNav} and \textit{ERPArm} policy under three input modalities.}
\label{fig06}
\end{figure}

\subsection{Training Details}
\label{sec:4.2}
Both \emph{ERPNav} and \emph{ERPArm} policies were trained in three input modalities: \emph{RGB-Only}, \emph{Event-Only}, and \emph{RGB+Event}. For the unimodal variants, the corresponding input modality was excluded, and the input layer was adjusted to reflect the channel dimensions. All models were trained with identical hyperparameters and protocols to ensure fair and consistent evaluation. For \emph{ERPNav}, the transformer-based architecture from Fig.~\ref{fig05} was used, with the policy head defined in (Section.~\ref{sec:3.4}). All variants were trained using the Adam optimizer with a learning rate of \(2\times10^{-4}\) and L$2$ weight decay of \(1\times10^{-4}\). The loss function was mean squared error (MSE), targeting continuous predictions of normalized linear (\(x\)) and angular (\(z\)) velocities in the range \([-1,1]\). Training ran for $10$ epochs with a batch size of $32$, using early stopping with a patience of $2$ epochs based on validation loss.
For \emph{ERPArm}, the same backbone was paired with a deeper policy head to regress $6$-DoF pre-grasp poses. Training also employed the Adam optimizer, with a lower initial learning rate of \(1\times10^{-4}\). A \texttt{ReduceLROnPlateau} scheduler was used to adaptively reduce the learning rate by half if the validation loss did not improve by at least \(1\times10^{-4}\) over two epochs. The Smooth L$1$ (Huber) Loss was used to regress the target pose vector. As with \textit{ERPNav}, training spanned $10$ epochs with batch size $32$ and early stopping.

\begin{table}[t]
\centering
\caption{ERPNav performance for tracking in simulation trials}
\label{table04}
\resizebox{0.80\columnwidth}{!}{%
\begin{tabular}{c|c|c|c|c}
\toprule 
\textbf{\begin{tabular}[c]{@{}c@{}}Policy\\ (Input Modality)\end{tabular}} &
  \textbf{\begin{tabular}[c]{@{}c@{}}Centroid Error\\ (px, $\mu \pm \sigma$)\end{tabular}} &
  \textbf{\begin{tabular}[c]{@{}c@{}}Mean BBox\\ Width (px)\end{tabular}} &
  \textbf{\begin{tabular}[c]{@{}c@{}}Success Rate\\ (\% within threshold)\end{tabular}} &
  \textbf{\begin{tabular}[c]{@{}c@{}}Mean Trial\\ Duration (s)\end{tabular}} \\ \midrule
\textbf{RGB}         & 129.7 $\pm$ 45.0          & \textbf{251.8} & 86.7          & 31.0          \\ \hline
\textbf{Event}       & 152.0 $\pm$ 44.0          & 175.7          & 73.3          & \textbf{41.6} \\ \hline
\textbf{RGB+Event} & \textbf{106.7 $\pm$ 26.3} & 196.6          & \textbf{93.3} & 35.3          \\ \bottomrule
\end{tabular}%
}
\end{table}

\begin{table}[t]
\centering
\caption{ERPArm performance under single and multi-object conditions}
\label{table05}
\resizebox{0.80\columnwidth}{!}{%
\begin{tabular}{c|c|c|c|c|c}
\toprule
\textbf{Scenario} &
  \textbf{\begin{tabular}[c]{@{}c@{}}Policy\\ (Input Modality)\end{tabular}} &
  \textbf{\begin{tabular}[c]{@{}c@{}}Error\\ (mm, $\mu \pm \sigma$)\end{tabular}} &
  \textbf{\begin{tabular}[c]{@{}c@{}}Accuracy\\ (\%)\end{tabular}} &
  \textbf{\begin{tabular}[c]{@{}c@{}}Latency\\ (ms, $\mu \pm\sigma$)\end{tabular}}&
  \textbf{\begin{tabular}[c]{@{}c@{}}Success Rate\\ (\%)\end{tabular}} \\ \midrule
\multirow{3}{*}{Single Object} 
 & RGB         & 48.6 $\pm$ 12.2 & 43.5 & 5.7 $\pm$  0.6 & 28.4 \\ \cline{2-6} 
 & Event       & 66.2 $\pm$  16.4 & 31.8 & \textbf{3.2 $\pm$  0.4} & 26.9 \\ \cline{2-6} 
 & RGB+Event &  \textbf{41.1 $\pm$  9.5}  & \textbf{71.4} & 7.8 $\pm$  0.6 & \textbf{51.7 }\\ \hline
\multirow{3}{*}{Multi Object}  
 & RGB         & 61.9 $\pm$  14.6 & 33.7 & 5.4 $\pm$  1.1 & 26.1 \\ \cline{2-6} 
 & Event       & 82.3 $\pm$  19.2 & 22.4  & \textbf{3.0 $\pm$  0.9} & 21.5 \\ \cline{2-6} 
 & RGB+Event & \textbf{52.6 $\pm$  11.3} & \textbf{58.9} & 7.6 $\pm$  0.5 & \textbf{31.8} \\ \bottomrule
\end{tabular}%
}
\end{table}

Fig.~\ref{fig06} shows the training and validation loss curves for both policies. For \textit{ERPNav}, the \emph{RGB+Event} variant consistently achieves the lowest loss, followed by \emph{RGB-Only}, then \emph{Event-Only}, which also converges more slowly. Similar trends are observed for \textit{ERPArm}, where fusing RGB and event data yields the best performance.
The quantitative results are shown in Table~\ref{table04} and~\ref{table05}. For \textit{ERPNav} (Table~\ref{table04}), the \emph{RGB+Event} model achieves the lowest mean centroid error and the highest success rate, indicating superior tracking stability. It means the bounding-box width also suggests optimal following distance without excessive closeness. The \textit{Event-only} model exhibits longer episode durations, suggesting slower convergence. 
For \textit{ERPArm} (Table~\ref{table05}), the \textit{RGB+Event} variant shows the lowest position error and highest accuracy and success rate under both single- and multi-object scenarios. Although the \textit{Event-only} variant offers the lowest latency, it underperforms in terms of accuracy and success rate, confirming the advantage of fusing modalities in event-driven grasp prediction.

\subsection{Simulation Results and Inference}
\label{sec:4.3}
The trained \textit{ERPNav} and \textit{ERPArm} policy variants were evaluated in the Gazebo simulator across $15$ episodes each, using inference nodes. Performance was assessed using task-specific metrics to compare the \emph{RGB-Only}, \emph{Event-Only}, and \emph{RGB+Event} models.
Table~\ref{table04} presents the evaluation metrics for the navigation task. Four metrics were used: ($1$) \textit{centroid error} (mean $\pm$ standard deviation), measuring the Euclidean distance between the target object’s centroid and the image center-lower values indicate better alignment, ($2$) \textit{bounding-box width}, representing the perceived proximity to the object larger widths suggest closer following, ($3$) \textit{success rate}, defined as the percentage of frames where the object remained within a 200 px radius of the image center, and ($4$) \textit{trial duration}, indicating average time per episode.
Among the three variants, the \emph{RGB+Event} policy achieved the lowest mean centroid error and smallest variance, reflecting stable and accurate tracking. It maintained a moderate bounding-box width, suggesting optimal proximity without risk of collision, and achieved the highest success rate, demonstrating superior robustness across trials.

Table~\ref{table05} summarizes results for single and multi-object grasping scenarios. In the single-object setting, the policy predicted a pre-grasp pose for a solitary cube on a conveyor belt. In the multi-object case, the pose was predicted for the leftmost object. Metrics include: ($1$) \textit{position error} (mm) between predicted and expert poses, ($2$) \textit{accuracy}, defined as the proportion of predictions within a set error threshold, ($3$) \textit{inference latency} (ms), and ($4$) \textit{success rate}, i.e., the percentage of successful grasps using predicted poses.
Across both scenarios, the \emph{RGB+Event} variant outperformed the other models, achieving the lowest error rates, highest grasp accuracy, and superior success rates, while maintaining inference latency within an acceptable range. These results underscore the benefit of early fusion of event and RGB modalities, especially in dynamic grasping tasks where precise spatial-temporal cues are essential.

\section{Conclusion}
\label{sec:conclusion}
This work introduced \textit{SEBVS}, a synthetic event-based visual servoing framework that fuses RGB and asynchronous event streams within a unified transformer architecture for robot navigation and manipulation. Through \textit{ERPNav} and \textit{ERPArm} policies, we demonstrated that combining high-resolution frames with microsecond event cues improves control accuracy, robustness, and task success over single-modality approaches.
In simulation, the \emph{RGB+Event} variant consistently outperformed both \emph{RGB-only} and \emph{Event-only} policies across diverse evaluation metrics. \textit{ERPNav} achieved lower centroid tracking error and higher success rates in dynamic tracking, while \textit{ERPArm} yielded more accurate pre-grasp pose predictions and higher grasp success under both single and multi-object settings. These results underscore the value of event data in enhancing perception and decision-making, especially in visually challenging or fast-moving environments. Importantly, this study also demonstrated the practicality of using event emulators like v2e within simulators such as Gazebo to train event-driven policies, enabling low-latency inference without the need for specialized neuromorphic hardware during development. Ongoing efforts aim to deploy \textit{SEBVS} on real-world robot platforms equipped with physical event cameras. To bridge the sim-to-real gap \cite{chanda2025event}, we will explore domain adaptation strategies, including domain randomization and fine-tuning. Additional directions include extending \textit{ERP} policies to cluttered scenes, multi-agent systems, and integrating temporal attention to better model event dynamics.

\bibliographystyle{splncs04}   
\bibliography{main}

\end{document}